\documentclass[runningheads]{llncs}

\usepackage[caption=false,font=footnotesize]{subfig}
\usepackage{graphicx}
\usepackage{booktabs}
\usepackage{url}
\usepackage{amsmath}
\usepackage{amssymb}
\usepackage{algorithm}
\usepackage{algpseudocode}
\usepackage{listings}
\usepackage{xspace}
\usepackage{fnpct}
\usepackage{tikz}
\usepackage{tabularx}
\newcolumntype{R}{>{\raggedleft\arraybackslash}X}

\usepackage{pgf,pgfplots,pgfplotstable,forest}
\usepgfplotslibrary{external}
\usetikzlibrary{positioning}
\usetikzlibrary{pgfplots.statistics,arrows,backgrounds}
\pgfplotsset{compat=1.14}

\newcommand{\z}{\ensuremath{\mathbf{z}}\xspace}
\newcommand{\fz}{\ensuremath{f_{\z}}\xspace}
\newcommand{\zopt}{\ensuremath{\z_{opt}}\xspace}

\newcommand{\lb}[1]{\ensuremath{\underline{\mathbf{#1}}}\xspace}
\newcommand{\ub}[1]{\ensuremath{\overline{\mathbf{#1}}}\xspace}
\newcommand{\estzub}{\ensuremath{\hat{\ub{\z}}}\xspace}
\newcommand{\estzlb}{\ensuremath{\hat{\lb{\z}}}\xspace}

\newcommand{\gtba}{\ensuremath{\text{GTB}_{a}}\xspace}
\newcommand{\gtbs}{\ensuremath{\text{GTB}_{s}}\xspace}
\newcommand{\nna}{\ensuremath{\text{NN}_{a}}\xspace}
\newcommand{\nns}{\ensuremath{\text{NN}_{s}}\xspace}

\newcommand{\method}{Bion\xspace}

\begin{document}

\title{Learning Objective Boundaries for Constraint Optimization Problems}

\author{Helge Spieker%
\and Arnaud Gotlieb%
}
\institute{Simula Research Laboratory, P.O. Box 134, 1325 Lysaker, Norway \email{\{helge,arnaud\}@simula.no}}

\maketitle

\begin{abstract}
Constraint Optimization Problems (COP) are often considered without sufficient knowledge on the boundaries of the objective variable to optimize. When available, tight boundaries are helpful to prune the search space or estimate problem characteristics. Finding close boundaries, that correctly under- and overestimate the optimum, is almost impossible without actually solving the COP. This paper introduces Bion, a novel approach for boundary estimation by learning from previously solved instances of the COP. Based on supervised machine learning, Bion is problem-specific and solver-independent and can be applied to any COP which is repeatedly solved with different data inputs. An experimental evaluation over seven realistic COPs shows that an estimation model can be trained to prune the objective variables’ domains by over 80\%. By evaluating the estimated boundaries with various COP solvers, we find that Bion improves the solving process for some problems, although the effect of closer bounds is generally problem-dependent.

\keywords{Machine Learning \and Constraint Optimization \and Objective Boundaries}
\end{abstract}

\section{Introduction}

Many scheduling or planning problems involve the exact optimization of some variable (e.g., timespan), that depends on decision variables, constrained by a set of combinatorial relations. 
These problems, called Constraint Optimization Problems (COP), are notoriously difficult to solve \cite{Hooker2012}. 
They are often addressed with systematic tree-search, such as branch-and-bound, where parts of the search space with worse cost than the current best solution are pruned. 
In Constraint Programming, these systematic techniques work without prior knowledge and are steered by the constraint model. 
Unfortunately, the worst-case computational cost to fully explore the search space is exponential in the worst case and the performance of the solver depends on efficient domain pruning \cite{Rossi2006}.

COP solving with branch-and-bound is often considered without sufficient knowledge on boundaries of the objective variable \cite{Milano2006}. When available, these boundaries can support the solver in discovering near-optimal solutions early during search and thus can reduce the computational effort \cite{Beck2011,Hooker2012,Gualandi2012}. In addition, providing tight estimates of the objective variable is useful user-feedback to estimate interesting features of COP \cite{DeRaedt2011}. 
Unfortunately, finding close under- and over estimations of the objective variable is still an open problem~\cite{Ha2015} and almost impossible without actually running the solver with a good heuristic. Domain boundaries are therefore usually obtained through problem-specific heuristics \cite{Beck2011,Gualandi2012,Liu2018}, which requires a deep understanding of the COP. Finding a generic method for closer domain estimation would allow many COP instances to be solved more efficiently.

This paper introduces \method, a new method combining logic-driven constraint optimization and supervised machine learning (ML), for solving COP. Using the known results of already-solved instances of a COP, a trained data-driven ML estimation model predicts boundaries for the objective variable of a new instance. These boundaries are then exploited by a COP solver to prune the search space. In simpler words, for a given COP, an estimation model is trained once with already solved instances of the same problem. For a new instance, the trained model is exploited to estimate close boundaries of the objective variable. Using the estimated boundaries, additional domain constraints are added to the COP model and used by the solver to prune the search space at low cost. 
Note however that ML methods can only approximate the optimum and are therefore not a full alternative. To eliminate the inherent risk of misestimations, which can cause unsatisfiability, the ML model is trained with an asymmetric loss function, adjusted training labels, and other counter-measures. 
As a result, \method is an exact method for solving COP and can complement advantageously any existing COP solver. Interestingly, the main computational cost of \method lies in the training part, but the estimation cost for a new input is low. 

Besides the general ability to estimate close objective boundaries, we explore with \method how useful these boundaries are to prune the search space and to improve the COP solving process. 
Our results show that boundary estimation can generally improve solver performance, even though there are dependencies on the right combination of solver and COP model for best use of the reduced domains. %
The main contributions of this paper are threefold:
\begin{enumerate}
\item We introduce \method, a new exact method combining ML and traditional COP solving to estimate close boundaries of the objective variable and to exploit these boundaries for boosting the solving process. \method can be advantageously applied to any COP which is repeatedly solved with different data inputs. To the best of our knowledge, this is the first time a problem- and solver-independent ML method using historical data is proposed.
\item We discuss training techniques to avoid misestimations, compare various ML models such as gradient tree boosting, support vector machine and neural networks, with symmetric and asymmetric loss functions. A contribution lies in the dedicated feature selection and user-parameterized label shift, a new method to train these models on COP characteristics.
\item We evaluate \method's ability to prune objective domains, as well as the impact of estimated and manually determined boundaries on solver performance with seven COPs.
\end{enumerate}

\section{Related Work}
\label{sec:related}

Many exact solvers include heuristics to initial feasible solutions that can be used for bounding the search, using for example, linear relaxations ~\cite{Hooker2012}. Others rely on branching heuristics and constraint propagation to find close bounds early~\cite{Rossi2006}. Recent works have also considered including the objective variable as part of the search and branch heuristic \cite{Fages2017,Palmieri2018}.
By exploiting a trained ML model, our approach \method adds an additional bounding step before the solver execution but it does not replace the solvers' bounding mechanisms. Hence, it complements these approaches by starting with a smaller search space.

The combination of ML and exact solvers has been previously explored from different angles~\cite{DeRaedt2011,Lombardi2018}. One angle is the usage of ML for solver configuration and algorithm selection, for example by selecting and configuring a search strategy~\cite{Arbelaez2010,Loth2013,Chu2015}, deciding when to run heuristics~\cite{Khalil2017} or lazy learning~\cite{Gent2010}, or to efficiently orchestrate a solver portfolio~\cite{Seipp2015,Amadini2015}. In \cite{Cappart2019}, Cappart \textit{et al.} propose the integration of reinforcement learning into the construction of decision diagrams to derive bounds for solutions of optimization problems. A framework for mutual integration of CP and ML was presented, the Inductive Constraint Programming (ICP) loop~\cite{Bessiere2017}, which is also applicable to \method.
Lombardi \textit{et al.} present a general framework for embedding ML-trained models in optimization techniques, called empirical model learning \cite{Lombardi2017}. This approach deploys trained ML models directly in the COP as additional global constraints, which is a promising approach, especially
when the learning model works directly on the model input. In our case, the feature set uses external instance descriptions and works on different input sizes, which is a different type of integration.
The coupling of data-driven and exact methods differs from the work on learning combinatorial optimization purely from training \cite{Bello2017,Dai2017,Deudon2018}, without using any constraint solver. As the complexity of a full solution is much higher than estimating the objective value, these methods require considerably more training data and computational resources, and are not yet competitive in practice.

Previous work also exists on predicting instance characteristics of COPs. These characteristics include runtime prediction, e.g., for the traveling salesperson problem~\cite{Mersmann2013,Hoos2014} or combining multiple
heuristics into a single admissible A* heuristic~\cite{Samadi2008}. However, these methods are tailored to a single problem and rely on problem-specific features for the statistical model, which makes them effective for their specific use-case. They require substantial analytic effort per problem. In contrast, our approach is both solver- and problem-independent. 
\section{Background}
\label{sec:background}

This section introduces constraint optimization in the context of Constraint Programming over Finite Domains~\cite{Rossi2006} and necessities of supervised ML.

We define a Constraint Optimization Problem (COP) as a triple $\langle \mathcal{X},\mathcal{C},\fz \rangle$ where
$\mathcal{X} = \{\mathbf{x}_1, \dots, \mathbf{x}_n\}$ is a set of variables, $\mathcal{C}$ is a set of constraints $\mathcal{C}=\{c_1,\dots,c_m\}$ and $\fz$ is an objective function with value \z to optimize. 
Each variable $\mathbf{x}_i$, also called decision variable, is associated with a finite domain $\mathcal{D}(\mathbf{x}_i)$, representing all possible values. 
A constraint $c_i$ is defined over a subset of $r_i$ variables, represented by all allowed $r_i$-tuples of the corresponding relation.  %

A (feasible) solution $\varphi$ of the COP is an assignment of each variable to a single value from its domain, such that all constraints are satisfied. Each solution corresponds to an objective value $\z_{\varphi} = \fz(\varphi)$.
The goal of solving the COP is to find at least one $\varphi^*$, called an optimal solution, such that $\fz(\varphi^*) = \zopt$ is optimal.
Depending on the problem formulation, \fz has either to be minimized or maximized by finding solutions with a smaller respectively larger objective value $\z$.
We use \lb{x} and \ub{x} to refer to the lower and upper domain boundaries of a
variable $\mathbf{x}$, and the notation $\lb{x}..\ub{x}$ to denote the domain, i.e. the integer
set $\{\,n \mid \lb{x} \leq n \leq \ub{x}\,\}$.

A COP $\langle \mathcal{X},\mathcal{C},\fz \rangle$ is \emph{satisfiable} iff it has at least one feasible solution. 
The instance is \emph{unsatisfiable} iff it has no feasible solution.
A COP is said to be \emph{solved} iff at least one of its feasible solution is proved optimal. 

A ML model can be trained for the regression task to approximate the objective function of a COP.
In a typical regression task, a continuous value $\hat{y}$ is predicted for a given input vector $\mathbf{x}$: $f(\mathbf{x}) = \hat{y}$, e.g., for predicting closer domain boundaries. %
The model can be trained through supervised learning, that is, by examples of input vectors $\mathbf{x}$ and true outputs $y$: $\{(\mathbf{x_1}, y_1), (\mathbf{x_2}, y_2), \dots, (\mathbf{x_m}, y_m)\}$.
During successive iterations, the model parameters are adjusted with the help of a loss function, that evaluates the approximation of training examples~\cite{Hastie2009}.

We now introduce the concept of estimated boundaries, 
which refers to providing close lower and upper bounds
for the optimal value \zopt.%
An \emph{estimation} is a domain $\estzlb..\estzub$ which defines boundaries
for the domain of $\fz$.
The domain boundaries are predicted by supervised ML, that is, $\estzub = f(\mathbf{x})$, $\estzlb = f(\mathbf{x})$.
    An estimation $\estzlb..\estzub$ is \emph{admissible} iff $\zopt \in \estzlb..\estzub$. %
    Otherwise, the estimation is \emph{inadmissible}.

We further classify the two domain boundaries as \emph{cutting} and
\emph{limiting} boundaries in relation to their effect on the solver's search process.
Depending on whether the COP is a minimization or maximization problem, these
terms refer to different domain boundaries.
The \emph{cutting boundary} is the domain boundary that reduces the number of reachable solutions. 
For minimization, this is the upper domain boundary \ub{\z}; 
for maximization, the lower domain boundary \lb{\z}.
Similarly, the \emph{limiting boundary} is the domain boundary that does not reduce the number of reachable solutions, but only reduces the search space to be explored.
For minimization, this is the lower domain boundary \lb{\z}; 
for maximization, the upper domain boundary \ub{\z}.
 
\section{Learning to Estimate Boundaries}
\label{sec:method_ml}

In this section, we explain the initial part of our method \method, which is to train a ML model to estimate close objective domain boundaries. This includes a discussion on the features to describe COPs and their instances, and how to train the model such that the estimated boundaries do not render the problem unsatisfiable or exclude the optimal objective value.

Training an estimator model has to be performed only once per COP.
From each instance of an existing dataset, a feature vector is extracted in order to train an estimator model. This model predicts both lower and upper boundaries for the objective variable of each instance.
When the estimated boundaries are used to support COP solving, they are embedded into the COP, either through additional hard constraints or with an augmented search strategy, which we will discuss in Section~\ref{sec:method_cp}. 

The training set is constructed such that the label $y$ of each COP instance, i.e., the true optimal objective value \z, is scaled by the original objective domain boundaries into $[0,1]$, with $\lb{z} \triangleq 0$ and $\ub{z} \triangleq 1$. After estimating the boundaries, the model output is scaled back from $[0,1]$ to the original domain.
This scaling allows the model to relate the approximated objective function to the initial domain. This is useful both if the given boundaries tend to systematically under- and overestimate the optimal objective, and as it steers the estimations to be within the original boundaries and therefore improve admissibility. Furthermore, some ML models, such as those based on neural networks, benefit from having an output within a specified range for their estimation performance.

\subsection{Instance Representation}
\label{sec:features}
The estimator ML model expects a fixed-size numeric feature vectors as its input.
This feature vector is calculated from the COP model and the instance by analyzing its values and the model structure.
As the presented method is problem-independent, a generic set of features is chosen, and problem-specific information is gathered by exploiting the variables and structure of the COP instance. 
This means, the set of features to describe a COP instance can be calculated for any instance of any COP model, which makes the method easily transferable without having to identify domain-dependent, specific features. 
Still, it is possible to extend the feature set by problem- or model-specific features to further improve the model performance.

At the same time, by requirement of the used ML methods, the size of the feature vector is fixed for all inputs of an estimator, i.e. for one COP model and all its possible instances.
However, in practice, the size of each COP instance varies and the instance cannot directly function as a feature vector. 
For example, the number of locations varies in routing problems, or the number of tasks varies in scheduling problems, with each having a couple of additional attributes per location or task.

We construct the generic feature vector from two main sets of features. 
The first set of features focuses on the instance parameters and their values, i.e. these features directly encode the input to the COP model.
The second set of features stem from the combination of COP model and instance and describe the resulting constraint system.

From the description of each decision variable of the COP, a first set of features is constructed. Thereby, each COP uses a problem-specific feature vector, depending on the number and the type of decision variables, constructed from problem-independent characteristics.
Each decision variable of the COP instance is processed and described individually. 
Finite domains variables and constants are directly used without further processing.
Data structures for collections of values, such as lists or sets, are described by $9$ measures from descriptive statistics. 
Multidimensional and nested data structures are aggregated to a single, fixed-size dimension by the sum of the features for each nested dimension.
The $9$ statistical measures are:
1) The number of values in the collection and their 2) minimum and 3) maximum; 4) standard deviation and 5) interquartile range for dispersion; 6) mean and 7) median for central tendency; 8) skew and 9) kurtosis to describe the distribution's shape.

Additionally, the second set of features characterize the complete constraint system of COP model and instance. 
These features analyze the number of variables, constraints, etc. for the COP~\cite{Amadini2014} and have originally been developed to efficiently schedule a portfolio of solvers~\cite{Amadini2015}. 
Some of these features are without relevant information, because they can be constant for all instances of a COP, e.g. the number of global constraints in the model. 
Although these features are less descriptive than the first set, we observed a small accuracy improvement from including them.

Finally, when the feature vectors for all instances in the training set have been constructed, a final step is to remove features with a low variance, that add little or no information, to reduce the number of inputs and model complexity.

\subsection{Eliminating Inadmissible Estimations}
\label{sec:avoidinfeasible}

The boundaries of a domain regulate which optimum values a variable can take. For the objective variable, this means which objective values can be reached. By further limiting the domain through an external estimation, the desired effect is to prune unnecessary values, such as low-quality objective values. This pruning focuses the search on the high-quality region of the objective space.

The trained estimator only approximates an instance’s objective value, but does not calculate it precisely, i.e., even after training estimation errors occur. 
These errors are usually approximately evenly distributed between under- and over estimations, because many applications make no difference in the error type.
However, in the boundary estimation case, the type of error is crucial. If the estimator underestimates, resp. overestimates, the objective value in a minimization, resp. maximization COP, all solutions, including the optima, are excluded from the resulting objective domain.
On the other hand, errors shifting the boundary away from the optimum only have the penalty of a larger resulting domain, but still find high-quality solutions, including the optima.

We consider three techniques to avoid inadmissible estimations. Two of these techniques, label shift and asymmetric losses, are applied during the estimator’s training phase. The third affects the integration of estimated boundaries during constraint solving and will be discussed in Section \ref{sec:method_cp}.

\subsubsection{Adjusting Training Labels}
\label{sec:labelshift}

As inadmissible estimations are costly, but some estimation error is acceptable, the first technique \textit{label shift} changes the training label, i.e. the objective value of the sample instances, by a small margin, such that a small estimation error still leads to an admissible estimation. 
The estimator is thereby trained to always under- or overestimate the objective value.
Label shift is similar to the prediction shift concept from \cite{Tolstikov2017}, but is specifically designed for boundary estimation and the application on COP models. 

Formally, label shift is defined as:
\begin{align*}
y' &= y + \lambda \,(\ub{z} - y)\,\quad \text{\emph{(Overestimation)}}\\
y' &= y - \lambda \,(y - \lb{z})\,\quad \text{\emph{(Underestimation)}}
\end{align*}
with adjustment factor $\lambda \in [0,1)$.
The configuration parameter $\lambda$ steers the label shift margin, with a smaller $\lambda$ being closer to the true label $y$. 
Therefore, setting $\lambda$ is a trade-off between close estimations and their feasibility.

\subsubsection{Training with Asymmetric Loss Functions}
\label{sec:asymmetricloss}

\begin{figure}[t]
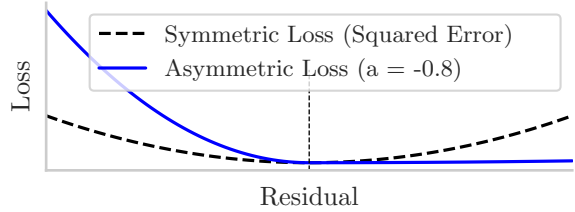

  \centering
\begingroup%
\makeatletter%
\begin{pgfpicture}%
\pgfpathrectangle{\pgfpointorigin}{\pgfqpoint{2.979679in}{1.103342in}}%
\pgfusepath{use as bounding box, clip}%
\begin{pgfscope}%
\pgfsetbuttcap%
\pgfsetmiterjoin%
\definecolor{currentfill}{rgb}{1.000000,1.000000,1.000000}%
\pgfsetfillcolor{currentfill}%
\pgfsetlinewidth{0.000000pt}%
\definecolor{currentstroke}{rgb}{1.000000,1.000000,1.000000}%
\pgfsetstrokecolor{currentstroke}%
\pgfsetdash{}{0pt}%
\pgfpathmoveto{\pgfqpoint{0.000000in}{0.000000in}}%
\pgfpathlineto{\pgfqpoint{2.979679in}{0.000000in}}%
\pgfpathlineto{\pgfqpoint{2.979679in}{1.103342in}}%
\pgfpathlineto{\pgfqpoint{0.000000in}{1.103342in}}%
\pgfpathclose%
\pgfusepath{fill}%
\end{pgfscope}%
\begin{pgfscope}%
\pgfsetbuttcap%
\pgfsetmiterjoin%
\definecolor{currentfill}{rgb}{1.000000,1.000000,1.000000}%
\pgfsetfillcolor{currentfill}%
\pgfsetlinewidth{0.000000pt}%
\definecolor{currentstroke}{rgb}{0.000000,0.000000,0.000000}%
\pgfsetstrokecolor{currentstroke}%
\pgfsetstrokeopacity{0.000000}%
\pgfsetdash{}{0pt}%
\pgfpathmoveto{\pgfqpoint{0.216536in}{0.216536in}}%
\pgfpathlineto{\pgfqpoint{2.970079in}{0.216536in}}%
\pgfpathlineto{\pgfqpoint{2.970079in}{1.093742in}}%
\pgfpathlineto{\pgfqpoint{0.216536in}{1.093742in}}%
\pgfpathclose%
\pgfusepath{fill}%
\end{pgfscope}%
\begin{pgfscope}%
\definecolor{textcolor}{rgb}{0.150000,0.150000,0.150000}%
\pgfsetstrokecolor{textcolor}%
\pgfsetfillcolor{textcolor}%
\pgftext[x=1.593308in,y=0.129036in,,top]{\color{textcolor}\rmfamily\fontsize{9.600000}{11.520000}\selectfont Residual}%
\end{pgfscope}%
\begin{pgfscope}%
\definecolor{textcolor}{rgb}{0.150000,0.150000,0.150000}%
\pgfsetstrokecolor{textcolor}%
\pgfsetfillcolor{textcolor}%
\pgftext[x=0.129036in,y=0.655139in,,bottom,rotate=90.000000]{\color{textcolor}\rmfamily\fontsize{9.600000}{11.520000}\selectfont Loss}%
\end{pgfscope}%
\begin{pgfscope}%
\pgfpathrectangle{\pgfqpoint{0.216536in}{0.216536in}}{\pgfqpoint{2.753543in}{0.877205in}}%
\pgfusepath{clip}%
\pgfsetbuttcap%
\pgfsetroundjoin%
\pgfsetlinewidth{1.204500pt}%
\definecolor{currentstroke}{rgb}{0.000000,0.000000,0.000000}%
\pgfsetstrokecolor{currentstroke}%
\pgfsetdash{{4.440000pt}{1.920000pt}}{0.000000pt}%
\pgfpathmoveto{\pgfqpoint{0.216536in}{0.502539in}}%
\pgfpathlineto{\pgfqpoint{0.262582in}{0.486351in}}%
\pgfpathlineto{\pgfqpoint{0.308628in}{0.470713in}}%
\pgfpathlineto{\pgfqpoint{0.354674in}{0.455626in}}%
\pgfpathlineto{\pgfqpoint{0.400720in}{0.441090in}}%
\pgfpathlineto{\pgfqpoint{0.446766in}{0.427104in}}%
\pgfpathlineto{\pgfqpoint{0.492812in}{0.413669in}}%
\pgfpathlineto{\pgfqpoint{0.538857in}{0.400784in}}%
\pgfpathlineto{\pgfqpoint{0.584903in}{0.388450in}}%
\pgfpathlineto{\pgfqpoint{0.630949in}{0.376667in}}%
\pgfpathlineto{\pgfqpoint{0.676995in}{0.365435in}}%
\pgfpathlineto{\pgfqpoint{0.723041in}{0.354753in}}%
\pgfpathlineto{\pgfqpoint{0.769087in}{0.344621in}}%
\pgfpathlineto{\pgfqpoint{0.815133in}{0.335040in}}%
\pgfpathlineto{\pgfqpoint{0.861179in}{0.326010in}}%
\pgfpathlineto{\pgfqpoint{0.907224in}{0.317531in}}%
\pgfpathlineto{\pgfqpoint{0.953270in}{0.309602in}}%
\pgfpathlineto{\pgfqpoint{0.999316in}{0.302224in}}%
\pgfpathlineto{\pgfqpoint{1.045362in}{0.295396in}}%
\pgfpathlineto{\pgfqpoint{1.091408in}{0.289119in}}%
\pgfpathlineto{\pgfqpoint{1.137454in}{0.283392in}}%
\pgfpathlineto{\pgfqpoint{1.183500in}{0.278217in}}%
\pgfpathlineto{\pgfqpoint{1.229546in}{0.273591in}}%
\pgfpathlineto{\pgfqpoint{1.275591in}{0.269517in}}%
\pgfpathlineto{\pgfqpoint{1.321637in}{0.265993in}}%
\pgfpathlineto{\pgfqpoint{1.367683in}{0.263019in}}%
\pgfpathlineto{\pgfqpoint{1.413729in}{0.260597in}}%
\pgfpathlineto{\pgfqpoint{1.459775in}{0.258725in}}%
\pgfpathlineto{\pgfqpoint{1.505821in}{0.257403in}}%
\pgfpathlineto{\pgfqpoint{1.551867in}{0.256632in}}%
\pgfpathlineto{\pgfqpoint{1.597912in}{0.256412in}}%
\pgfpathlineto{\pgfqpoint{1.643958in}{0.256742in}}%
\pgfpathlineto{\pgfqpoint{1.690004in}{0.257623in}}%
\pgfpathlineto{\pgfqpoint{1.736050in}{0.259055in}}%
\pgfpathlineto{\pgfqpoint{1.782096in}{0.261037in}}%
\pgfpathlineto{\pgfqpoint{1.828142in}{0.263570in}}%
\pgfpathlineto{\pgfqpoint{1.874188in}{0.266654in}}%
\pgfpathlineto{\pgfqpoint{1.920234in}{0.270288in}}%
\pgfpathlineto{\pgfqpoint{1.966279in}{0.274472in}}%
\pgfpathlineto{\pgfqpoint{2.012325in}{0.279208in}}%
\pgfpathlineto{\pgfqpoint{2.058371in}{0.284494in}}%
\pgfpathlineto{\pgfqpoint{2.104417in}{0.290330in}}%
\pgfpathlineto{\pgfqpoint{2.150463in}{0.296717in}}%
\pgfpathlineto{\pgfqpoint{2.196509in}{0.303655in}}%
\pgfpathlineto{\pgfqpoint{2.242555in}{0.311144in}}%
\pgfpathlineto{\pgfqpoint{2.288601in}{0.319183in}}%
\pgfpathlineto{\pgfqpoint{2.334646in}{0.327772in}}%
\pgfpathlineto{\pgfqpoint{2.380692in}{0.336913in}}%
\pgfpathlineto{\pgfqpoint{2.426738in}{0.346603in}}%
\pgfpathlineto{\pgfqpoint{2.472784in}{0.356845in}}%
\pgfpathlineto{\pgfqpoint{2.518830in}{0.367637in}}%
\pgfpathlineto{\pgfqpoint{2.564876in}{0.378980in}}%
\pgfpathlineto{\pgfqpoint{2.610922in}{0.390873in}}%
\pgfpathlineto{\pgfqpoint{2.656968in}{0.403317in}}%
\pgfpathlineto{\pgfqpoint{2.703013in}{0.416312in}}%
\pgfpathlineto{\pgfqpoint{2.749059in}{0.429857in}}%
\pgfpathlineto{\pgfqpoint{2.795105in}{0.443953in}}%
\pgfpathlineto{\pgfqpoint{2.841151in}{0.458599in}}%
\pgfpathlineto{\pgfqpoint{2.887197in}{0.473796in}}%
\pgfpathlineto{\pgfqpoint{2.933243in}{0.489544in}}%
\pgfpathlineto{\pgfqpoint{2.970079in}{0.502539in}}%
\pgfpathlineto{\pgfqpoint{2.970079in}{0.502539in}}%
\pgfusepath{stroke}%
\end{pgfscope}%
\begin{pgfscope}%
\pgfpathrectangle{\pgfqpoint{0.216536in}{0.216536in}}{\pgfqpoint{2.753543in}{0.877205in}}%
\pgfusepath{clip}%
\pgfsetroundcap%
\pgfsetroundjoin%
\pgfsetlinewidth{1.204500pt}%
\definecolor{currentstroke}{rgb}{0.000000,0.000000,1.000000}%
\pgfsetstrokecolor{currentstroke}%
\pgfsetdash{}{0pt}%
\pgfpathmoveto{\pgfqpoint{0.216536in}{1.053869in}}%
\pgfpathlineto{\pgfqpoint{0.244164in}{1.022185in}}%
\pgfpathlineto{\pgfqpoint{0.271791in}{0.991143in}}%
\pgfpathlineto{\pgfqpoint{0.299419in}{0.960744in}}%
\pgfpathlineto{\pgfqpoint{0.327046in}{0.930987in}}%
\pgfpathlineto{\pgfqpoint{0.354674in}{0.901872in}}%
\pgfpathlineto{\pgfqpoint{0.382302in}{0.873399in}}%
\pgfpathlineto{\pgfqpoint{0.409929in}{0.845568in}}%
\pgfpathlineto{\pgfqpoint{0.437557in}{0.818380in}}%
\pgfpathlineto{\pgfqpoint{0.465184in}{0.791834in}}%
\pgfpathlineto{\pgfqpoint{0.492812in}{0.765930in}}%
\pgfpathlineto{\pgfqpoint{0.520439in}{0.740669in}}%
\pgfpathlineto{\pgfqpoint{0.548067in}{0.716050in}}%
\pgfpathlineto{\pgfqpoint{0.575694in}{0.692072in}}%
\pgfpathlineto{\pgfqpoint{0.603322in}{0.668738in}}%
\pgfpathlineto{\pgfqpoint{0.630949in}{0.646045in}}%
\pgfpathlineto{\pgfqpoint{0.658577in}{0.623995in}}%
\pgfpathlineto{\pgfqpoint{0.686204in}{0.602587in}}%
\pgfpathlineto{\pgfqpoint{0.713832in}{0.581821in}}%
\pgfpathlineto{\pgfqpoint{0.741459in}{0.561697in}}%
\pgfpathlineto{\pgfqpoint{0.769087in}{0.542216in}}%
\pgfpathlineto{\pgfqpoint{0.796714in}{0.523377in}}%
\pgfpathlineto{\pgfqpoint{0.824342in}{0.505180in}}%
\pgfpathlineto{\pgfqpoint{0.851969in}{0.487625in}}%
\pgfpathlineto{\pgfqpoint{0.879597in}{0.470713in}}%
\pgfpathlineto{\pgfqpoint{0.907224in}{0.454443in}}%
\pgfpathlineto{\pgfqpoint{0.934852in}{0.438815in}}%
\pgfpathlineto{\pgfqpoint{0.962479in}{0.423829in}}%
\pgfpathlineto{\pgfqpoint{0.990107in}{0.409486in}}%
\pgfpathlineto{\pgfqpoint{1.017735in}{0.395785in}}%
\pgfpathlineto{\pgfqpoint{1.045362in}{0.382726in}}%
\pgfpathlineto{\pgfqpoint{1.072990in}{0.370309in}}%
\pgfpathlineto{\pgfqpoint{1.100617in}{0.358535in}}%
\pgfpathlineto{\pgfqpoint{1.128245in}{0.347402in}}%
\pgfpathlineto{\pgfqpoint{1.155872in}{0.336913in}}%
\pgfpathlineto{\pgfqpoint{1.183500in}{0.327065in}}%
\pgfpathlineto{\pgfqpoint{1.211127in}{0.317859in}}%
\pgfpathlineto{\pgfqpoint{1.238755in}{0.309296in}}%
\pgfpathlineto{\pgfqpoint{1.266382in}{0.301375in}}%
\pgfpathlineto{\pgfqpoint{1.294010in}{0.294096in}}%
\pgfpathlineto{\pgfqpoint{1.321637in}{0.287460in}}%
\pgfpathlineto{\pgfqpoint{1.349265in}{0.281466in}}%
\pgfpathlineto{\pgfqpoint{1.376892in}{0.276114in}}%
\pgfpathlineto{\pgfqpoint{1.404520in}{0.271404in}}%
\pgfpathlineto{\pgfqpoint{1.432147in}{0.267336in}}%
\pgfpathlineto{\pgfqpoint{1.459775in}{0.263911in}}%
\pgfpathlineto{\pgfqpoint{1.487402in}{0.261128in}}%
\pgfpathlineto{\pgfqpoint{1.515030in}{0.258987in}}%
\pgfpathlineto{\pgfqpoint{1.542657in}{0.257489in}}%
\pgfpathlineto{\pgfqpoint{1.570285in}{0.256632in}}%
\pgfpathlineto{\pgfqpoint{1.597912in}{0.256409in}}%
\pgfpathlineto{\pgfqpoint{1.800514in}{0.256632in}}%
\pgfpathlineto{\pgfqpoint{2.003116in}{0.257282in}}%
\pgfpathlineto{\pgfqpoint{2.205718in}{0.258357in}}%
\pgfpathlineto{\pgfqpoint{2.408320in}{0.259859in}}%
\pgfpathlineto{\pgfqpoint{2.610922in}{0.261788in}}%
\pgfpathlineto{\pgfqpoint{2.813523in}{0.264143in}}%
\pgfpathlineto{\pgfqpoint{2.970079in}{0.266254in}}%
\pgfpathlineto{\pgfqpoint{2.970079in}{0.266254in}}%
\pgfusepath{stroke}%
\end{pgfscope}%
\begin{pgfscope}%
\pgfpathrectangle{\pgfqpoint{0.216536in}{0.216536in}}{\pgfqpoint{2.753543in}{0.877205in}}%
\pgfusepath{clip}%
\pgfsetbuttcap%
\pgfsetroundjoin%
\pgfsetlinewidth{0.501875pt}%
\definecolor{currentstroke}{rgb}{0.000000,0.000000,0.000000}%
\pgfsetstrokecolor{currentstroke}%
\pgfsetdash{{1.850000pt}{0.800000pt}}{0.000000pt}%
\pgfpathmoveto{\pgfqpoint{1.593308in}{0.216536in}}%
\pgfpathlineto{\pgfqpoint{1.593308in}{0.742860in}}%
\pgfusepath{stroke}%
\end{pgfscope}%
\begin{pgfscope}%
\pgfsetrectcap%
\pgfsetmiterjoin%
\pgfsetlinewidth{1.003750pt}%
\definecolor{currentstroke}{rgb}{0.800000,0.800000,0.800000}%
\pgfsetstrokecolor{currentstroke}%
\pgfsetdash{}{0pt}%
\pgfpathmoveto{\pgfqpoint{0.216536in}{0.216536in}}%
\pgfpathlineto{\pgfqpoint{0.216536in}{1.093742in}}%
\pgfusepath{stroke}%
\end{pgfscope}%
\begin{pgfscope}%
\pgfsetrectcap%
\pgfsetmiterjoin%
\pgfsetlinewidth{1.003750pt}%
\definecolor{currentstroke}{rgb}{0.800000,0.800000,0.800000}%
\pgfsetstrokecolor{currentstroke}%
\pgfsetdash{}{0pt}%
\pgfpathmoveto{\pgfqpoint{0.216536in}{0.216536in}}%
\pgfpathlineto{\pgfqpoint{2.970079in}{0.216536in}}%
\pgfusepath{stroke}%
\end{pgfscope}%
\begin{pgfscope}%
\pgfsetbuttcap%
\pgfsetmiterjoin%
\definecolor{currentfill}{rgb}{1.000000,1.000000,1.000000}%
\pgfsetfillcolor{currentfill}%
\pgfsetfillopacity{0.800000}%
\pgfsetlinewidth{0.803000pt}%
\definecolor{currentstroke}{rgb}{0.800000,0.800000,0.800000}%
\pgfsetstrokecolor{currentstroke}%
\pgfsetstrokeopacity{0.800000}%
\pgfsetdash{}{0pt}%
\pgfpathmoveto{\pgfqpoint{0.468688in}{0.633714in}}%
\pgfpathlineto{\pgfqpoint{2.884524in}{0.633714in}}%
\pgfpathquadraticcurveto{\pgfqpoint{2.908968in}{0.633714in}}{\pgfqpoint{2.908968in}{0.658159in}}%
\pgfpathlineto{\pgfqpoint{2.908968in}{1.008186in}}%
\pgfpathquadraticcurveto{\pgfqpoint{2.908968in}{1.032631in}}{\pgfqpoint{2.884524in}{1.032631in}}%
\pgfpathlineto{\pgfqpoint{0.468688in}{1.032631in}}%
\pgfpathquadraticcurveto{\pgfqpoint{0.444244in}{1.032631in}}{\pgfqpoint{0.444244in}{1.008186in}}%
\pgfpathlineto{\pgfqpoint{0.444244in}{0.658159in}}%
\pgfpathquadraticcurveto{\pgfqpoint{0.444244in}{0.633714in}}{\pgfqpoint{0.468688in}{0.633714in}}%
\pgfpathclose%
\pgfusepath{stroke,fill}%
\end{pgfscope}%
\begin{pgfscope}%
\pgfsetbuttcap%
\pgfsetroundjoin%
\pgfsetlinewidth{1.204500pt}%
\definecolor{currentstroke}{rgb}{0.000000,0.000000,0.000000}%
\pgfsetstrokecolor{currentstroke}%
\pgfsetdash{{4.440000pt}{1.920000pt}}{0.000000pt}%
\pgfpathmoveto{\pgfqpoint{0.493133in}{0.933659in}}%
\pgfpathlineto{\pgfqpoint{0.737577in}{0.933659in}}%
\pgfusepath{stroke}%
\end{pgfscope}%
\begin{pgfscope}%
\definecolor{textcolor}{rgb}{0.150000,0.150000,0.150000}%
\pgfsetstrokecolor{textcolor}%
\pgfsetfillcolor{textcolor}%
\pgftext[x=0.835355in,y=0.890882in,left,base]{\color{textcolor}\rmfamily\fontsize{8.800000}{10.560000}\selectfont Symmetric Loss (Squared Error)}%
\end{pgfscope}%
\begin{pgfscope}%
\pgfsetroundcap%
\pgfsetroundjoin%
\pgfsetlinewidth{1.204500pt}%
\definecolor{currentstroke}{rgb}{0.000000,0.000000,1.000000}%
\pgfsetstrokecolor{currentstroke}%
\pgfsetdash{}{0pt}%
\pgfpathmoveto{\pgfqpoint{0.493133in}{0.752534in}}%
\pgfpathlineto{\pgfqpoint{0.737577in}{0.752534in}}%
\pgfusepath{stroke}%
\end{pgfscope}%
\begin{pgfscope}%
\definecolor{textcolor}{rgb}{0.150000,0.150000,0.150000}%
\pgfsetstrokecolor{textcolor}%
\pgfsetfillcolor{textcolor}%
\pgftext[x=0.835355in,y=0.709757in,left,base]{\color{textcolor}\rmfamily\fontsize{8.800000}{10.560000}\selectfont Asymmetric Loss (a = -0.8)}%
\end{pgfscope}%
\end{pgfpicture}%
\makeatother%
\endgroup%
   \caption{Symmetric and asymmetric loss functions. The asymmetric loss assigns
    a higher loss to a negative residuals, but lower loss to overestimations.}
  \label{fig:loss_functions}
\end{figure}

ML models are usually trained with symmetric loss functions, that do not differentiate between positive and negative errors. An asymmetric loss function, on the other hand, assigns higher loss values for either under- or overestimations, which penalizes certain errors stronger than others. Figure~\ref{fig:loss_functions} shows an example of quadratic symmetric and asymmetric loss functions and the difference in penalization.

\emph{Shifted Squared Error Loss} is an imbalanced variant of squared error loss. Formally speaking,
the shifted squared error loss is defined as
\begin{align*}
L(r) = r^2 \cdot (sgn(r) + \alpha)^2 \text{ with absolute error } r = \hat{y} - y
\end{align*}
where $\hat{y}$ is the estimated value and $y$ is the true target value. 
The parameter $\alpha$ shifts the penalization towards under- or overestimation and influences the magnitude of the penalty.

\subsection{Estimated Boundaries during Search}  %
\label{sec:method_cp}

One potential application of estimated objective boundaries is their usage to improve the COP solving process.
Using \method to solve a COP consists of the following steps:
1) (Initially) Train an estimator model for the COP;
    2) Extract a feature vector from each COP instance;
    3) Estimate both a lower and an upper objective boundaries;
    4) Update the COP with estimated boundaries; and 
    5) Solve the updated COP with the solver.

The boundaries provided by the estimator can be embedded as hard constraints on the objective variable, i.e., by adding $\z \in \estzlb \dots \estzub$.
The induced overhead is negligible, but dealing with misestimations requires additional control.
If all feasible solutions are excluded, because the cutting bound is wrongly estimated, the instance is rendered unsatisfiable.
This issue is handled by reverting to the original domain.
If only optimal solutions are excluded, because the limiting bound is wrongly estimated, then only non-optimal solutions can be returned and this stays impossible to notice.
This issue cannot be detected in a single optimization run of the solver. 
However, in practical cases where the goal is to find good-enough solutions early rather than finding truly-proven optima, it can be an acceptable risk to come-up with an good approximation of the optimal solutions only.
In conclusion, hard boundary constraints are especially suited for cases where a high confidence in the quality of the estimator has been gained, and the occurrence of inadmissible estimations is unlikely.

\section{Experimental Evaluation}
\label{sec:experiments}
We evaluate our method in three experiments, which focus 1) on the impact of label shift and asymmetric loss functions for training the estimator, 2) on the estimators' performance to bound the objective domain, and 3) on the impact of estimated boundaries on solver performance.

We selected the seven COP having the most instances from the MiniZinc benchmark repository\footnote{\url{github.com/MiniZinc/minizinc-benchmarks}}, some of which are deployed in MiniZinc challenges \cite{Stuckey2014}.
These seven COPs are MRCPSP (11182 instances), RCPSP (2904 inst.), Bin Packing (500 inst.), Cutting Stock (121 inst.), Jobshop (74 inst.), VRP (74 inst.), and Open Stacks (50 inst.). 
Considering training sets of different sizes, from 50 to over 11,000 instances, is relevant to understand scenarios that can benefit from boundary estimation.

We consider four ML models to estimate boundaries: neural networks (NN), gradient tree boosting (GTB), support vector machines (SVM) and linear regression (LR). NN and GTB come in two variants, using either symmetric (\gtbs, \nns) or asymmetric (\gtba, \nna) loss functions. 
All models are used with their default hyperparameters as defined by the libraries. The NN is a feed-forward neural network with $5$ layers of $64$ hidden nodes, a larger NN did not show to improve the results. \method is implemented in Python, using scikit-learn~\cite{Pedregosa2011} for SVM and LR. To support the implementation of asymmetric loss functions, NNs are based on Keras~\cite{Chollet2015}, and GTB on XGBoost~\cite{Chen2016}.

Our experiments are focused towards the general
effectiveness of \method over a range of problems. 
Therefore, we used the default parameters and did not perform parameter tuning, although it could improve the performance.
As loss factors for the asymmetric loss functions, we set \(a = -1\) for \gtba and \(a = -0.8\) for \nna, where a smaller \(a\) caused problems during training. 

We trained the ML models on commodity hardware without GPU acceleration, and the training time took less than $5$ sec. per model, except for MRCPSP with up to $6$ minutes with the NN model. 
The estimation performance of the ML models is measured via repeated 10-fold validation.
The training set is split randomly into 10 folds, of which 9 folds are used to train the model and 1 fold for evaluation.
This step is repeated 10 times, so each part is used for evaluation once.
We report the median results over 10 runs.

\subsection{Avoiding Inadmissible Estimations}
The first experiment focuses on label shift, a technique to avoid inadmissible estimations. Label shift changes the training label of the ML model towards under- or overestimation based on an adjustment factor $\lambda$.
Setting $\lambda$ is a trade-off between close boundaries and inadmissible estimations. For the experiment, we considered $11 \lambda$ values between $0$ and $0.8$ for every estimator variant.

All models benefit from training with label shift, but the adjustment factor for best performance varies. 
The asymmetric models \nna and \gtba only require small $\lambda = 0.1$, respectively 0.3, to maximize the number of admissible estimations. 
For LR ($\lambda = 0.5$), SVM ($0.8$), \nns ($0.5$), and \gtbs ($0.4$), label shift is necessary to reach a large amount of admissible estimations. 
Here, the optimal $\lambda$ is approximately $0.5$, which shifts the label in the middle between optimum and domain boundary. 
As symmetric models do not distinguish between under- and overestimation error, this confirms the trade-off characteristic of $\lambda$. 
These results underline the benefit of training with both label shift and asymmetric loss functions for avoiding inadmissible estimations.

\subsection{Estimating Tighter Domain Boundaries}

\begin{table*}[t]
  \centering
\begin{tabularx}{\textwidth}{lRR|RR|RR|RR|RR|RR}
\toprule
{} & \multicolumn{2}{c}{GTB$_a$} & \multicolumn{2}{c}{GTB$_s$} & \multicolumn{2}{c}{LR} & \multicolumn{2}{c}{NN$_a$} & \multicolumn{2}{c}{NN$_s$} & \multicolumn{2}{c}{SVM} \\
{} &    Gap &       Size &    Gap &      Size &     Gap &      Size &      Gap &       Size &    Gap &       Size &    Gap &      Size \\
\midrule
Bin Packing   &   68 &   65 &   60 &  58 &    48 &  48 &    \textbf{78}\textsuperscript{*} &  \textbf{68}\textsuperscript{*} &   50 &   48 &   15 &  18 \\
Cutting Stock &  \textbf{64}\textsuperscript{*} &   66 &  58\textsuperscript{*} &  59 &   48\textsuperscript{*} &  49 &  41\textsuperscript{***} &  \textbf{71}\textsuperscript{+} &  48\textsuperscript{*} &   49 &  29\textsuperscript{*} &  17 \\
Jobshop   &   69 &   69 &   60 &  60 &    50 &  50 & \textbf{87} &   \textbf{81} &   50 &   48 &   19 &  20 \\
MRCPSP    &   64 &   61 &   60 &  59 &    49 &  49 & \textbf{80} &   \textbf{76} &   49 &   49 &   13 &  19 \\
Open Stacks   &  \textbf{64}\textsuperscript{*} &  \textbf{60}\textsuperscript{+} &  59\textsuperscript{*} &  53 &  43\textsuperscript{**} &  43 &   56\textsuperscript{**} &  33\textsuperscript{*} &  47\textsuperscript{*} &  42\textsuperscript{+} &  15\textsuperscript{+} &  15 \\
RCPSP     &   65 &   64 &   60 &  60 &    50 &  50 &    \textbf{80}\textsuperscript{+} &  \textbf{76}\textsuperscript{+} &   50 &   50 &   13 &  20 \\
VRP   &   70 &   70 &   60 &  60 &    50 &  50 & \textbf{89} &   \textbf{88} &   50 &   50 &    0 &   0 \\
\bottomrule
\end{tabularx}
   \caption{Reduction in objective domain through estimated boundaries (in \%). 
  \emph{Gap}: Domain size between cutting boundary and optimum ($(1 - (|\estzub - \zopt|/|\ub{\z} - \zopt|)) * 100$).
  \emph{Size}: Ratio between new and initial domain size ($(1 - (|\estzub - \estzlb|/|\ub{\z} - \lb{\z}|)) * 100$).
  Cells show the median and the median absolute deviation (MAD): No superscript indicator \(\leq
  5 \leq {}^{+} \leq 10 < {}^{*} \leq 20 < {}^{**} \leq 30 < {}^{***}\). 
    }
  \label{tab:prediction}
\end{table*}

We analyze here the capability of each model to estimate tight domain boundaries, as compared to the original domains of the COP. As evaluation metrics, the size of the estimated domain is compared to the original domain size. Furthermore, the distance between cutting boundary and optimal objective value is compared between the estimated and original domain. A closer gap between cutting bound and objective value leads to a relatively better first solution when using the estimations and is therefore of practical interest.
Table \ref{tab:prediction} shows the estimation performance per problem and estimator.
The results show that asymmetric models are able to estimate closer boundaries than symmetric models. For each model, the estimation performance is consistent over all problems. 

First, we look at the share of admissible estimations. Most models achieve $100$ \% admissible estimations in all problems. Exceptions exist for Cutting Stock (\gtba , \gtbs, LR: 91 \%, SVM: 50 \%) and RCPSP (\nns, SVM: 83 \%, all other models: $\ge 98 \%$). In general, \nna has the highest number of admissible estimations, followed by \gtba. 
The largest reduction is achieved by \nna, 
making it the overall best performing model.
\gtba is also capable to consistently reduce the domain size by over $60$ \%, but not as much as \nna.
Cutting Stock and Open Stacks are difficult problems for most models, as indicated by the deviations in the results.
LR and \nns reduce the domain size by approximately 50 \%, when the label shift adjustment factor $\lambda$ is $0.5$, as selected in the previous experiment.

Conclusively, these results show that \method has an excellent ability to derive general estimation models from the extracted instance features. The estimators reduce substantially the domains and provide tight boundaries.

\subsection{Effects on Solver Performance}
Our third experiment investigated the effect of objective boundaries on the solver performance. 
The setup for the experiments is as follows.
For each COP, 30 instances were randomly selected.
Each instance was run in four configurations, using: 
1) The original COP model without any modification;
2) The COP model with added upper and lower boundary constraints, estimated by \method with \nna;
3) The COP model with only an upper boundary constraint, estimated by \method;
4) The COP model with a user-fixed upper boundary, set up on the middle between the true optimum and the first found solution when using no boundary constraints ($\z_{first}$): $\ub{z} = \zopt + \lfloor(\z_{first} - \zopt) / 2\rfloor$. 
This model is a baseline to evaluate whether there is any benefit in solving the COP with additional boundary constraints.

\begin{table*}[t]
    \footnotesize
  \centering
  \subfloat[Chuffed]{%
\begin{tabularx}{\textwidth}{lRRR|RRR|RRR}
\toprule
{} & \multicolumn{3}{c}{Equiv. Solution Time} & \multicolumn{3}{c}{Quality of First} & \multicolumn{3}{c}{Time to Completion} \\
{} & Fixed &  Upper &  Both & Fixed & Upper &  Both & Fixed &   Upper &    Both \\
\midrule
Bin Packing   &  -9.4 &   36.1 &  13.0 & -37.9 & -57.7 & -57.7 &  36.1 &  2140.7 &  2364.5 \\
Jobshop       & -96.5 &  -96.4 & -96.6 & -38.1 & -60.0 & -60.0 & -27.6 &   -53.6 &   -42.5 \\
MRCPSP        &   0.0 &    0.0 &   0.0 & -10.8 &  -0.4 &   0.0 &   1.2 &     0.3 &    -3.4 \\
Open Stacks   &  -1.3 &   -1.3 &  -0.9 & -24.0 & -13.2 & -13.2 &   2.0 &    -0.4 &     2.9 \\
RCPSP         &  -3.2 &  197.4 &  25.3 &  -3.3 &   0.0 &   0.0 &  -4.2 &     0.0 &    -4.2 \\
VRP           &   0.4 &    0.0 &   0.0 & -23.5 &   0.0 &   0.0 &   2.0 &     7.0 &     7.0 \\
\bottomrule
\end{tabularx}
}\\
  \subfloat[Gecode]{%
\begin{tabularx}{\textwidth}{lRRR|RRR|RRR}
\toprule
Bin Packing   &    53.5 &   0.3 &  -0.6 &  -4.7 &   0.0 &   0.0 & -10.3 &  -4.0 & -13.0 \\
Cutting Stock &  5627.0 &   7.3 & -29.5 &  -8.5 &  -5.5 &  -2.6 &   --  &   --  &   -- \\
Jobshop       &   189.3 &  -6.4 &  37.4 & -10.9 &   6.1 &   6.1 &   0.0 &   0.0 &   0.0 \\
MRCPSP        &     0.0 &   0.0 &  23.6 & -10.8 &  -0.4 &  -0.2 &   1.3 &   0.0 &   4.0 \\
Open Stacks   &     0.0 &  -1.5 &   0.0 & -24.0 & -12.8 & -12.8 &   8.9 &   6.4 &   6.8 \\
RCPSP         &   -17.2 &  56.8 & -14.4 &  -2.8 &   0.0 &   0.0 & -11.8 &  12.0 &  -9.4 \\
VRP           &     0.0 &   0.0 &   0.0 & -21.0 &   0.0 &   0.0 & -19.0 & -18.0 &  -8.0 \\
\bottomrule
\end{tabularx}
}\\
  \subfloat[OR-Tools]{%
\begin{tabularx}{\textwidth}{lRRR|RRR|RRR}
\toprule
Bin Packing   & -22.7 &   35.0 &  39.2 & -37.4 & -57.0 & -57.2 &  104.4 &  170.0 &  172.4 \\
Jobshop       &   1.1 &    0.0 &   0.0 & -16.5 &  -0.8 &  -0.8 &    0.0 &    0.0 &    0.0 \\
MRCPSP        &  -3.2 &   -3.0 &  45.3 & -10.8 &  -0.4 &   0.0 &   -2.4 &   -2.1 &    1.2 \\
Open Stacks   &  -5.0 &   -2.6 &  -3.1 & -24.0 & -13.2 & -13.2 &    6.3 &   -1.2 &    2.3 \\
RCPSP         &   0.0 &  147.2 &  30.4 &  -3.3 &   0.0 &   0.0 &   -6.6 &   27.0 &    7.8 \\
VRP           & -95.3 &    0.0 &   0.0 & -38.2 &   0.0 &   0.0 &   32.0 &   -3.0 &   -5.0 \\
\bottomrule
\end{tabularx}
}
  \caption{Effect of boundaries on solver performance (in \%).
  \emph{Fixed}: Upper boundary set to middle between optimum and first found solution of unbounded run.
  \emph{Upper}: Upper boundary set to estimated boundary.
  \emph{Both}: Upper and lower boundary set to estimated boundaries.
  Results are averaged over 30 instances, lower values are better.%
  }
  \label{tab:solver}
\end{table*}

We selected three distinct State-of-the-Art COP solvers, among those which have the highest rank in MiniZinc challenges: Chuffed (as distributed with MiniZinc 2.1.7) \cite{Chu2016}, Gecode 6.0.1 \cite{Schulte2018}, and Google OR-Tools 6.8. 
All runs were performed with a $4$-hour timeout on a single-core of an Intel E5-2670 with 2.6GHz.

Three metrics were used for evaluation (all in \,\%), each comparing a run with added boundary constraint to the original COP model. The \emph{Equivalent Solution Time} compares the time taken by the original model to reach a solution of similar or better quality than the first found solution when using \method. It is calculated as
$(t_{Bounds} - t_{Original})/t_{Original} * 100$.
The \emph{Quality of First} compares the quality of the first found solutions with and without boundary constraints and is calculated as $(1 - \z_{Bounds} / \z_{Original}) * 100$.
The \emph{Time to Completion} relates the times until the search completed and the optimal solution is found. It is calculated in the same way as the Equivalent Solution Time.

The results are shown in Table~\ref{tab:solver}, listed per solver and problem. 
The results for the Cutting Stock problem for Chuffed and OR-Tools are not given, because none of the configurations, including the original COP, found a solution for more than one instance. Gecode found at least one solution for $26$ of $30$ instances.
We obtain mixed results for the different solvers and problems, which indicates that benefiting from objective boundaries is both problem- and solver-specific.
This holds true both for the boundaries determined by boundary estimation (columns \emph{Upper} and \emph{Both}) and the user-fixed boundary (column \emph{Fixed}).

The general intuition, also confirmed by the literature, is that in many cases a reduced solution space allows more efficient search and for several COPs, this is confirmed. An interpretation for why the boundary constraints in some cases hinder effective search, compared to the original COP, is that the solvers can apply different improvement techniques for domain pruning or search once an initial solution is found. 
The best results are obtained for solving Jobshop with Chuffed, where the constraints improve both the time to find a good initial solution and the time until the search is completed. Whether both an upper and lower boundary constraint can be useful is visible for the combination of Gecode and RCPSP. Here, posting only the upper boundary constraint is not beneficial for the Equivalent Solution Time, but with both upper and lower boundary Gecode is 14\,\% faster than without any boundaries. A similar behaviour shows for Chuffed and RCPSP regarding Time to Completion, where only the upper boundary has no effect, but posting both bounds reduces the total solving time by 4\,\%.
At the same time, we observe that posting both upper and lower boundaries, even though they reduce the original domain boundaries, does not always help the solver, such as for example in the combination of Chuffed and Jobshop. 
This can come from the efficiency of search heuristics, which can somethimes reach better initial solutions than those obtained with \method in some cases.

In conclusion, our method \method can generally produce objective variable boundaries which are helpful to improve the solver performance. Still, it is open to understand which combination of solvers, search heuristics and COP model benefits the most from strongly reduced domains. 
To the best of our knowledge, no clear answer is yet available in the literature.
From the comparison with user-fixed boundaries that are known to reduce the solution space, we observe that the estimated boundaries with \method are competitive and provide a similar behaviour. 
This makes \method a promising approach in many contexts where COP are solved without any prior knowledge on initial boundaries. 
 
\section{Conclusion}
\label{sec:conclusion}
This paper presents \method, a boundary estimation method for constraint optimization problems (COP), based on machine learning (ML). 
A supervised ML-model is trained with solved instances of a given COP, in order to estimate close boundaries of the objective variable. We utilize two training techniques, namely, asymmetric loss functions and label shift, and another counter-measure to adjust automatically the training labels, and discard any wrong estimation. \method is lightweight and both solver- and problem-independent. Our experimental results, obtained on $7$ realistic COPs, show that already a small set of instances is sufficient to train an estimator model to reduce the domain size by over 80\,\%.

Solving practical assignment, planning or scheduling problems often requires to repeatedly solve the same COP with different inputs \cite{Ernst2004,Szeredi2016,Mossige2017,Spieker2019}.
Our approach is especially well-suited for those scenarios, where training data can be collected from previous iterations and historical data.

In future work, we plan to explore in depth the actual effects of objective boundaries on constraint solver performances with the goal to better predict in which scenarios \method can be the most beneficial. Another research lead we intend to follow is to integrate \method with search heuristics that explicitly focus on evaluating the objective variable \cite{Fages2017,Palmieri2018}. 
\bibliographystyle{splncs04}
\bibliography{refs.bib}

\end{document}